\documentclass[a4paper, 10pt, conference]{ieeeconf}

\usepackage{graphicx}
\graphicspath{{figures/}}

\usepackage{booktabs}
\usepackage{array}
\usepackage{tabularx}
\usepackage{xcolor}
\usepackage{amsmath}
\usepackage{comment}
\usepackage{bm}
\graphicspath{{figures/}}
\usepackage{lscape}

\newenvironment{conditions}[1][where:]
  {#1 \begin{tabular}[t]{>{$}l<{$} @{${}={}$} l}}
  {\end{tabular}\\[\belowdisplayskip]}

\newcolumntype{s}{>{\hsize=1.97cm}X}

\IEEEoverridecommandlockouts                             
                                                         
\title{\LARGE \bf
Engagement Decision Support for Beyond Visual Range Air Combat}

\author{
Joao~P.~A.~Dantas$^{1}$, Andre~N.~Costa$^{1}$, Diego~Geraldo$^{1}$, Marcos~R.~O.~A.~Maximo$^{2}$, and Takashi~Yoneyama$^{3}$%
\thanks{$^{1}$Joao~P.~A.~Dantas, Andre~N.~Costa, and Diego~Geraldo
are with Institute for Advanced Studies, 
        Sao Jose dos Campos, SP, 12.288-001, Brazil
        {\tt\small\{dantasjpad,negraoanc,diegodg\}@fab.mil.br}}%
\thanks{$^{2}$Marcos~R.~O.~A.~Maximo is with Autonomous Computational System Lab (LAB-SCA), Computer Science Division, Aeronautics Institute of Technology, 
        Sao Jose dos Campos, SP, 12228-900, Brazil
        {\tt\small mmaximo@ita.br}}%
\thanks{$^{3}$Takashi~Yoneyama is with Electronic Engineering Division, Aeronautics Institute of Technology,
        Sao Jose dos Campos, SP, 12228-900, Brazil
        {\tt\small takashi@ita.br}}%
}

\begin{document}

\maketitle

\begin{abstract}

This work aims to provide an engagement decision support tool for Beyond Visual Range (BVR) air combat in the context of Defensive Counter Air (DCA) missions. In BVR air combat, engagement decision refers to the choice of the moment the pilot engages a target by assuming an offensive stance and executing corresponding maneuvers. To model this decision, we use the Brazilian Air Force's Aerospace Simulation Environment (\textit{Ambiente de Simulação Aeroespacial - ASA} in Portuguese), which generated 3,729 constructive simulations lasting 12 minutes each and a total of 10,316 engagements. We analyzed all samples by an operational metric called the DCA index, which represents, based on the experience of subject matter experts, the degree of success in this type of mission. This metric considers the distances of the aircraft of the same team and the opposite team, the point of Combat Air Patrol, and the number of missiles used. By defining the engagement status right before it starts and the average of the DCA index throughout the engagement, we create a supervised learning model to determine the quality of a new engagement. An algorithm based on decision trees, working with the XGBoost library, provides a regression model to predict the DCA index with a coefficient of determination close to 0.8 and a Root Mean Square Error of 0.05 that can furnish parameters to the BVR pilot to decide whether or not to engage. Thus, using data obtained through simulations, this work contributes by building a decision support system based on machine learning for BVR air combat.

\end{abstract}

\section{INTRODUCTION}
Air combat may occur in two primary forms: Within Visual Range (WVR) and Beyond Visual Range (BVR)~\cite{kurniawan2019empirical}, with the latter being recently more developed in the operational context, due to more extensive availability of more advanced weapons and sensors~\cite{higby2005promise}. Notice that, even though modern air combat may still end WVR, through a series of complex decisions and maneuvers, it usually begins BVR, which frequently is the most critical phase of the combat since it may provide advantages and drawbacks for succeeding phases~\cite{yuan2016design}. There is no clear definition of the distance to differentiate these two forms of air combat since this may be subject to the conditions in which the air combat happens.

BVR conditions force the pilots to rely more on a series of systems to compose their situational awareness, allowing them to make tactical decisions during the combat, such as whether to fire a missile or not. Especially in BVR combat, the missile launches and the circumstances around these events are critical~\cite{aronsson2019supporting}, since these weapons are the main form to engage the opponent. Since BVR combat is rarely observable in practice, with low availability of historical data, much of the assessment of its possibilities must be done through simulation~\cite{stillion2015trends}. It is also true due to the high costs of flying, air space regulations, and limited availability of platforms representative of those used by opposing forces~\cite{kallstrom2020agent}.

This work contributes by developing a decision support tool for pilots in BVR combat situations based on simulated data, with a particular focus on solving the problem of deciding when to engage a specific enemy aircraft, which is commonly based solely on pilot experience. There are several previous approaches that relate to this decision, mainly providing different forms of modeling pilot behavior. Many of them use game theory to model air combat~\cite{karelahti2006game,virtanen2006modeling,mukai2003sequential,ha2018stochastic}. Other approaches found in the literature are as follows: Bayesian Networks~\cite{poropudas2007analyzing,fu2021air,rao2011situation,du2010study}, fuzzy logic~\cite{akabari2005fuzzy,prabhu2014decision}, agent-based modeling~\cite{heinze1998thinking}, influence diagrams~\cite{lin2007sequential}, reinforcement learning~\cite{hu2021application,toubman2016rapid,piao2020beyond,weilin2018decision}, artificial neural networks~\cite{dantas2018,yao2021study}, evolutionary algorithms~\cite{li2020multi,yang2020evasive}, minimax method~\cite{kang2019beyond}, and behavior trees~\cite{yao2015adaptive}.

\begin{comment}
combat~\cite{karelahti2006game,virtanen2006modeling,mukai2003sequential,davidovitz1989two,ma2019cooperative,ha2018stochastic}. Other approaches found in the literature are as follows: Bayesian Networks~\cite{poropudas2007analyzing,fu2021air,rao2011situation,du2010study}, fuzzy logic~\cite{akabari2005fuzzy,tran2002adaptation,prabhu2014decision}, agent-based modeling~\cite{heinze1998thinking}, influence diagrams~\cite{lin2007sequential}, reinforcement learning~\cite{hu2021application,toubman2016rapid,piao2020beyond,weilin2018decision}, artificial neural networks~\cite{dantas2018,yao2021study}, evolutionary algorithms~\cite{li2020multi,yang2020evasive}, minimax method~\cite{kang2019beyond}, and behavior trees~\cite{yao2015adaptive}.
\end{comment}

Among all of these approaches, Ha \emph{et al.} (2018)~\cite{ha2018stochastic} is the one that more directly focuses on estimating the firing moment, which is done through a probabilistic function of the target's evasive maneuverability, the missile's speed on the final approach, and the accuracy of the target's information to guide the missile. However, as with many of the other cited methods, the methodology proposed in~\cite{ha2018stochastic} has not been tested in simulations with a higher degree of fidelity concerning an actual BVR air combat.

As an alternative, our work uses supervised machine learning models based on decision trees, using the XGBoost library~\cite{chen2015xgboost}, to provide pilots with parameters that improve their situational awareness in air combat from data collected from simulations of operational scenarios, allowing them to better decide when to engage a target. Compared to most of the approaches, this study was conducted through higher fidelity simulations, providing systems and subsystems that resemble their natural counterparts through our simulation environment, such as 6 degrees of freedom (6DOF) multi-role combat aircraft, electronic warfare (EW) devices, datalink communications, and active radar-guided missiles.

This study is, therefore, developed around four main subjects: i) determination of BVR air combat scenarios that will be analyzed, ii) carrying out simulations based on the chosen scenarios in the agents' configurations are varied, iii) collecting and analyzing the data generated in the simulations, and iv) using machine learning techniques on this data to improve the pilot's situational awareness by providing information about the analyzed air combat, fulfilling the role of a decision support system.

The remainder of this paper is organized as follows. In section 2, the main characteristics of the aircraft in the BVR air combat are described. We also explain the Defensive Counter Air (DCA) index, the operational metric used to perform the engagement analysis. Besides, the sampling process of the simulation inputs is discussed. In section 3, we show the methodology used to create the solution approach. Section 4 investigates the results and analysis concerning the exploratory data analysis and model results. Finally, in section 5, we present the conclusions and future works.

\section{PROPOSAL DESCRIPTION}

This section shows the Fighter Agent model, describing the simulation agent's reasoning that we propose. Next, the DCA index, the operational metric that we created to evaluate the air combat engagements, is presented. Finally, we discuss our process of sampling simulation input parameters.

\subsection{Fighter Agent}

The Institute for Advanced Studies from the Brazilian Air Force develops the Aerospace Simulation Environment (\emph{Ambiente de Simulação Aeroespacial - ASA} in Portuguese) to provide a computational solution that enables the simulation of operational scenarios, allowing users to establish scenarios, parameters, and command decisions to support the development of tactics, techniques, and procedures. Since there is so much variability in the nature of military scenarios, it may be hard to define what the scenarios of interest are. Thus, the development of ASA is not limited to predefined layouts; instead, it poses itself as a flexible solution that may be tailored to the user's needs. Its modularity contributes to this flexibility since one can configure components (models) in diverse ways and combine them freely, which enables the creation of the most varied scenarios~\cite{costa2019master}.

ASA is a custom-made, object-oriented (C++), and high-fidelity environment that generates the simulations and the data to analyze the problem. The simulation concept addressed in this context is one in which the scenario elements are represented as agents capable of making decisions based on artificial intelligence models or arbitrary rules previously stipulated. Simulations of this nature, termed constructive, can be used in the decision-making process, for example, to predict possible outcomes of engagements between opposing forces and assist in the definition of lines of action.

\begin{figure}
\centering
\includegraphics[width=0.45\textwidth]{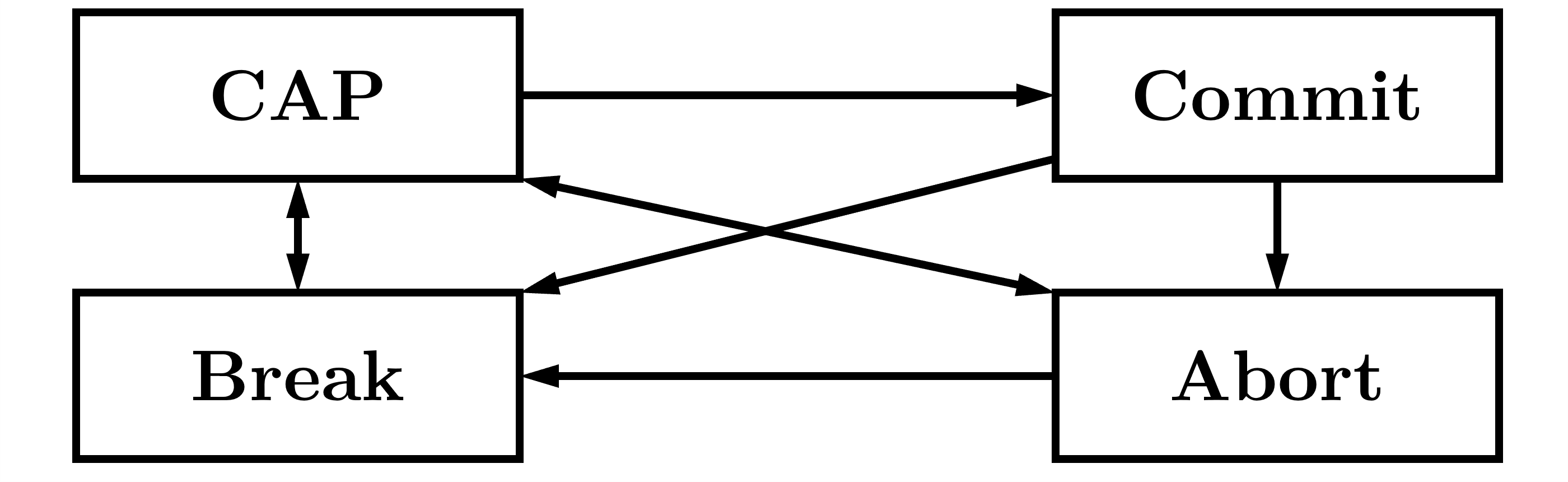}
\caption{FSM of agent tactics.} \label{fig1}
\end{figure}

The agent's actions are managed by a set of rules that describe a pilot's main tactical behaviors during a BVR combat. This set of rules is modeled as a Finite State Machine (FSM), or finite automaton, which is a mathematical computer model that, at any time, can be in only one state of a limited set of states~\cite{mamessier2014pilot}. Each state is one of the possible tactics that can be taken during a BVR combat. The following behaviors will be analyzed to better understand the presented problem: Combat Air Patrol (CAP), Commit, Abort, and Break. Fig.~\ref{fig1} shows the FSM that manages all the agent's tactics.

The CAP tactic consists of performing a flight pattern, describing an orbit, that can be any tactical maneuver, such as a circle or an oval, around a specific position called CAP point, with a defined heading and direction (clockwise or counterclockwise). The Commit tactic consists of the agent engaging a target detected by its radar or shared by data link by its allies. In the Abort tactic, the agent performs a defensive maneuver to move away from his priority threat, which is also the priority target in many situations. Finally, when the agent's sensors detect a missile threat fired in its direction, it performs the Break tactic. This tactic consists of a sudden defensive maneuver, describing a curve and a dive with great acceleration.

At the engagement moment, the FSM assumes the Commit behavior and will keep performing offensive maneuvers until the time it is desirable within combat. The offense and vulnerability indices are essential variables that guide the change from one state to another within the FSM. Then, when the agent needs to do a defensive movement, it assumes Break or Abort. Therefore, the engagement time in a simulation is defined as the time between the first Commit carried by the agent and its first Break or Abort. We created an operational index that determines the quality of the actions taken by the agent in a BVR scenario, which was calculated during the engagement period and can be extended to any moment in the simulation. This metric is referred to as the DCA Index, which will be detailed in the following subsection.

\subsection{DCA Index}

We defined the index as a probability of success, ranging from 0\% to 100\%, for BVR combat on DCA missions whose objective is to establish a CAP. These missions have the goal of defending a point of interest, which is done by ensuring that the opposing aircraft are kept far away from it. In addition, it aims to do that while launching the least number of missiles possible, which is interesting both from economic and operational standpoints. Furthermore, from the doctrine perspective, one may consider it good practice for the defending aircraft to stay close to its CAP point since it is easier to employ tactics to defend the point of interest.

\begin{figure}
\centering
\includegraphics[width=0.4\textwidth]{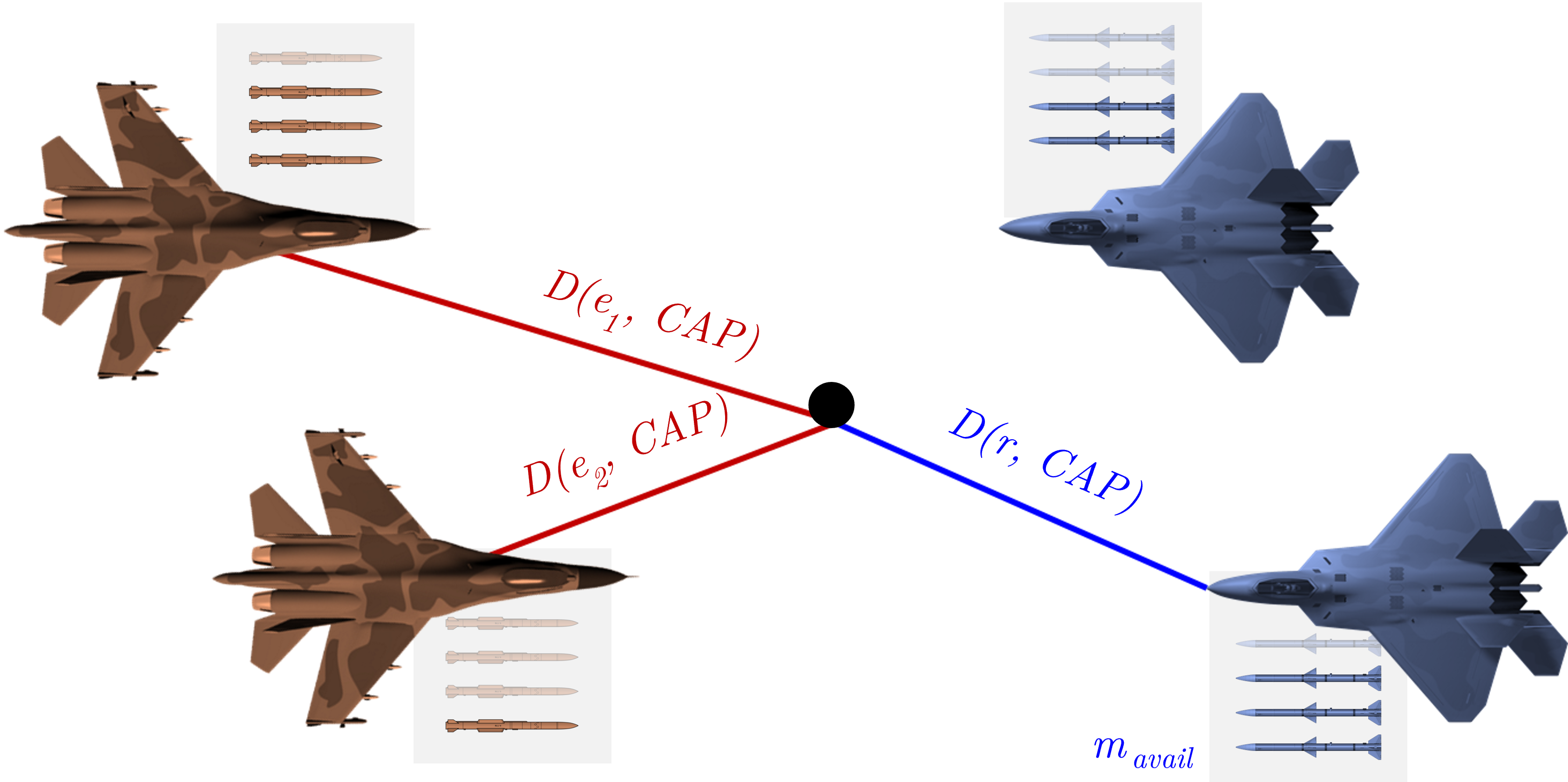}
\caption{Representation of the factors that form the DCA index.} \label{fig2}
\end{figure}

From these considerations with respect to the DCA mission context, we defined three basic principles (depicted in Fig.~\ref{fig2}) for the elaboration of the DCA index:
\begin{enumerate}
    \item Minimize the number of missile launched in the mission: ($m_{total}-m_{avail}$).
    \item Minimize the reference distance from its CAP point: $D(r,CAP)$.
    \item Maximize the distance of each enemy ($e_n$) to the CAP point: $D(e_{n},CAP)$.
\end{enumerate}

Firstly, at the beginning of the engagement, each aircraft has a fixed number of missiles ($n_{total}$), which at that moment is also the number of missiles available ($n_{avail}$). When the aircraft launches a missile, $n_{avail}$ decreases to keep track of the currently available missiles. The ratio between $n_{avail}$ and $n_{total}$ is one of the factors of the DCA index, so that, when maximizing this ratio, $n_{total}-n_{avail}$ is minimized.%as stated in~\eqref{eq1}.

Secondly, regarding minimizing the distance between the reference aircraft and the CAP point, i.e., $D(r,CAP)$, we considered that the decay of this effect is not linear since, as the distance increases, its influence becomes much less relevant to fulfilling the mission goals. Therefore, we chose a sigmoid function to generate such decay to encompass this non-linearity. We defined the sigmoid limits, considering operational experience with respect to the maximum range of the available missile, as 8,000 meters ($x_{99\%, r}$) for the 99\% output value, which corresponds to $y_{99\%, r}\approx4,5951$, and as 12,000 ($x_{1\%, r}$) meters for the 1\% output value, which stands for $y_{1\%, r}\approx-4,5951$. These limits must be used in a linear interpolation~\eqref{eq1} to convert the current distance value ($D_r$) to be input ($d_{r}$) in the sigmoid equation~\eqref{eq2}.

\begin{equation}\label{eq1}
d_i = \frac{(y_{99\%, i}-y_{1\%, i})}{(x_{99\%, i}-x_{1\%, i})} \cdot [D(i,CAP)-x_{1\%, i}] + y_{1\%, i}
\end{equation}
\begin{conditions}
i & $r$ (reference) or $e_n$ (enemy) \\
D_i &  measured distance from the CAP point \\
d_i &  interpolated distance for sigmoid input \\
\end{conditions}

Lastly, when considering the opposing aircraft distances ($D(e_{n},CAP)$), for all $N$ enemies, we used a similar sigmoid function, but with opposite characteristics since the idea was to increase the enemies' influence when they were closer to the CAP point. Therefore, the sigmoid limits for this factor were instead 12,000 ($x_{99\%, e}$) meters for the 99\% output value, which corresponds to $y_{99\%, e}\approx4,5951$, and 8,000 ($x_{99\%, e}$) meters for the 1\% output value, which stands for $y_{1\%, e}\approx-4,5951$. Applying the enemy sigmoid limits in the interpolation equation \eqref{eq1}, we are able to convert the measured distance value $D(e_n,CAP)$ to be input ($d_{e_n}$) in \eqref{eq2} for each of the enemy aircraft.

With the composition of the previously calculated factors, it is possible to obtain the DCA index's final calculation. The index factors have coefficients to allow prioritization of the three principles presented, as seen in~\eqref{eq2}. We defined the weights ($w_1=0.2$, $w_2=0.4$, and $w_3=0.4$) for each factor of the DCA index ($I_{DCA}$) based on the operational knowledge from subject matter experts.

\begin{equation}\label{eq2}
\begin{aligned}
I_{DCA} = &\; w_1 \cdot \frac{m_{avail}}{m_{total}} + w_2 \cdot \frac{1}{1+\exp(-d_r)} \\
          & + w_3 \cdot \frac{1}{N} \sum_{n=1}^N \frac{1}{1+\exp(-d_{e_n})}
\end{aligned}
\end{equation}

\subsection{Sampling of Simulation Input Parameters}

Latin Hypercube Sampling (LHS) is a method that can be used to produce, in a distributed way, the set of input values to be used in the simulations according to the desired intervals~\cite{wang2019formal}. This statistical method generates a random sample of parameter values from distribution in multiple dimensions. LHS consists of subdividing the sample universe into several disjoint subsets and extracting a representative element for each subsets, chosen at random.

The generated and stored simulations used in the supervised learning model were carried out in packages (batches). Thus, for the same type of scenario, it was possible to vary some parameters at the beginning of the simulation, which may lead to different outcomes. The simulations parameters changed during the sampling were: (a) latitude and longitude, determining the initial positions of the agents around the CAP points (adopting fixed CAP positions), (b) the flight level blocks to determine their altitudes, (c) the commit distance (the minimum distance that an agent is from a possible target that allows it to leave the CAP tactic and commit), (d) the thresholds of the offense and vulnerability indices before and after firing a missile (represent the level of risk acceptance that the agent is willing to withstand), (e) the shot philosophy (orientation defined during mission planning, before the flight, referring to the moment when, within the Weapon Engagement Zone (WEZ), which is an estimation of the missile maximum launch range~\cite{dantasbracis2021}, the agent must launch a missile), (f) the shot distance (the minimum distance that an agent is from a possible target that allows, during an engagement, fire a missile, i.e., the WEZ), and (g) the presence or absence on the aircraft of a Radar Warning Receiver (RWR), an EW system that detects electromagnetic emissions from opposing radar systems.

Using the LHS algorithm, 3,729 constructive simulations were generated in ASA, and a total of 10,316 engagements were observed. Each simulation corresponds to a 12-minute scenario executed three times faster than real-time, lasting approximately ten days in total.

The scenario consists of two opposing formations with two aircraft each, which are initially approaching, disengaged, and outside the radar range of each other. Their main goal is to establish a CAP at the same CAP point, invariably leading to a confrontation. When they enter the limits of the opponents' radars, the engagement phase begins. In the modeling proposed in this work, each aircraft is equipped with four of the same type of medium-range missile, i.e., up to 40 nautical miles in range. 

\section{METHODOLOGY}

After sampling the variables, respecting the intervals chosen for each one, we executed the simulations through ASA. The engagement events between the four agents are extracted with the data generated from these simulations. The average DCA index is calculated for all engagements extracted from the interval under analysis, generating the output variable that we intend to predict later for new samples. With the input data of the simulations and the output variable already defined, we build a supervised machine learning model based on eXtreme Gradient Boosting (XGBoost), predicting the average value of the agents' DCA index in future engagements. XGBoost represents a class of algorithms based on Decision Trees with Gradient Boosting~\cite{chen2016xgboost}. Its performance is analyzed with the test dataset after the model's training process is completed. This section discusses the input and output model variables, preprocessing procedures, hyperparameters tunning, evaluation metrics, and cross-validation processes.

\begin{comment}

\subsection{XGBoost}
XGBoost (eXtreme Gradient Boosting) is one of the most used algorithms by data scientists, presenting superior results, mainly in forecasting problems involving structured/tabular data~\cite{chen2016xgboost}. It represents a class of algorithms based on Decision Trees with Gradient Boosting, which means that the method uses the Gradient Descent algorithm to minimize loss while new models are being added. Due to its characteristics, the process can deal efficiently (and robustly) with a wide variety of data types. Furthermore, XGBoost is very flexible since it has many hyperparameters that can be improved, allowing it to be properly adjusted to different scenarios and problems. In fact, since its inception, it was considered to be the state of the art supervised machine learning algorithm for dealing with structured data~\cite{chen2015xgboost}. We decided to use XGBoost to analyze the described data in the proposed problem because of its speed, performance, and parallelization provision.
\end{comment}

\subsection{Model Input and Target Variables}
The main variables that coordinate the simulation of a BVR air combat were analyzed, and seventeen input variables were determined to be the most important to define the progress of the simulations based on the described scenarios. In addition, there are categorical and numerical variables with different ranges of coverage, and the definition of these sampling intervals was made based on the operational knowledge of BVR pilots and combat specialists. Next, in Table~\ref{tab2}, the description of each of the input variables of the simulations is carried out, presenting its unit when the variable is not dimensionless.

\begin{table}
\caption{Variables at the beginning of the engagement.}\label{tab2}
\begin{tabularx}{0.4875\textwidth}{| c| X|}

\hline      
\textbf{Parameter}   & \textbf{Description} \\        
\hline      
distance [m]  & Distance between the reference and the target \\
\hline      
aspect [deg] & Angle between the longitudinal axis of the target (projected rearward) and the line-of-sight to the reference \\
\hline      

delta\_head [deg] & Angle between the longitudinal axis of both aircraft \\
\hline      

delta\_alt [m]  & Difference of altitude between the reference and the target \\
\hline      
delta\_vel [kn]   & Difference of absolute velocity between the reference and the target  \\
\hline      
wez\_max\_o2t [m]  & Maximum range of the reference's weapon (non-maneuverable target) \\
\hline      
wez\_nez\_o2t [m]  & No-escape zone range of the reference's weapon (target performing high performance maneuver) \\
\hline      
wez\_max\_t2o [m]  & Estimated maximum range of the target's weapon (non-maneuverable reference) \\
\hline      
wez\_nez\_t2o [m]  & No-escape zone range of the target's weapon (reference performing high performance maneuver) \\
\hline      
vul\_thr\_bef\_shot       & Level of risk acceptance before shooting                     \\
\hline      
vul\_thr\_aft\_shot       & Level of risk acceptance after shooting                    \\
\hline      
shot\_point       & Missile firing point between the maximum range and the no-escape zone range of the reference                     \\
\hline      
rwr\_warning       & Boolean indicating whether the aircraft is equipped with an active RWR            \\
\hline      
hp\_tgt\_off       & High priority target offense index of the reference       \\
\hline      
hp\_thr\_vul       & High priority threat vulnerability index of the aircraft that is threatening the reference   \\
\hline      
own\_shot\_phi       & Reference shot philosophy                      \\
\hline      
enemy\_shot\_phi       & Estimated enemy's shot philosophy   \\ \hline
\end{tabularx}
\end{table}

Concerning the target parameter, the DCA index will be averaged between the start of the agent's Commit maneuver and the beginning of either Break or Abort maneuvers since both will make the agent disengage. Therefore, given a sequence of input parameters that define the agent's state, the model must predict the average value of the DCA index in that interval, improving the agent's situational awareness.

\subsection{Data Preprocessing}
Unlike what is done in artificial neural networks, there is no need to carry out data normalization procedures to employ the XGBoost algorithm since it is based on decision trees. Thus, using this type of learning method, the model benefits from one of the significant advantages of these trees in artificial intelligence problems related to the low amount of preprocessing required for it to be applied. It is necessary to transform the model's categorical variables into numerical ones to be appropriately processed. For this purpose, two preprocessing steps will be performed on the data: Label Encoding and One-Hot Encoding~\cite{cohen2013applied}. Also, feature engineering will be carried out to facilitate the training process of the proposed regression model.

\subsection{Hyperparameters Tuning}
GridSearch is a hyperparameter adjustment process to determine the ideal values for a given model, based on searching throughout a grid~\cite{putatunda2018comparative}. The performance of the entire model is based on the specified hyperparameter values. Some functions have been implemented, such as the GridSearchCV of the sklearn library, to automate finding the best of these values for the model. We performed an adjustment of the XGBoost model hyperparameters based on a variation of the library default values as observed in Table~\ref{tab3}.

\begin{table}[h]
\centering
\caption{GridSearch parameters for the prediction model.} \label{tab3}
\begin{tabular}{|c|c|}
\hline
\textbf{Parameters}  & \textbf{Values}                    \\
\hline
n\_estimators      & {[}100, 1000, 5000{]}             \\
learning\_rate     & {[}2, 3, 6, 10, 15, 20{]}         \\
max\_depth         & {[}0.1, 0.01, 0.001{]}            \\
gamma              & {[}0.0, 0.1, 0.2, 0.3, 0.4, 0.5{]} \\
subsample          & {[}0.6, 0.7, 0.8, 0.9{]}          \\
colsample\_bytree  & {[}0.6, 0.7, 0.8, 0.9{]}          \\
reg\_alpha         & {[}0.001, 0.01, 0.1, 1, 10, 100{]} \\
min\_child\_weight & {[}1,3,5,7,9,10,13,15{]}           \\ \hline
\end{tabular}
\end{table}

\subsection{Metrics}

The Root-Mean-Square Error (RMSE) is used to measure the differences between the values predicted by a model or an estimator concerning the actual values, providing the average magnitude of the error. As the errors are squared, the RMSE places a relatively high weight on significant errors, which means that RMSE should be most useful when large errors are particularly undesirable. In addition, the advantage of using the RMSE is that it has the same magnitude as the target variable, which helps in the interpretation of the average of the model errors found. In the analysis of the predictive models carried out in this work, the coefficient of determination ($R^2$) will also be used to evaluate the best architectures of the supervised machine learning models for BVR combat modeling. The use of $R^2$ is well established in classical regression analysis and it generally describes how the model is adapted to reality when making predictions.

\subsection{Cross-Validation}

% After performing a dataset train-test split, allocating 80\% for the training and 20\% for the testing, we propose to conduct cross-validation to evaluate the model's ability to predict new data that was not used in the estimate, to highlight problems such as overfitting~\cite{hyndman2006another}. A form of addressing this problem is through k-fold cross-validation, which benefits from using the whole train dataset for training and validation~\cite{willmott2006use}. We chose $k = 10$ after considering the trade-off between processing time and the generalization of the model results.
After performing a dataset train-validation-test split, allocating 80\% for the training and validation and 20\% for the testing, we propose to conduct cross-validation to evaluate the model's ability to predict new data that was not used in the estimate with the benefits of using the whole train dataset for training and validation and to highlight problems such as overfitting~\cite{hyndman2006another}.  We employ 10-fold cross-validation to address this problem after considering the trade-off between processing time and the generalization of the model results.
\section{RESULTS AND ANALYSIS}

This section presents a whole dataset exploratory data analysis to provide an initial understanding of the variables in the model, followed by 10-fold cross-validation results.

\subsection{Exploratory Data Analysis}

\begin{figure}[h]
\centering

\includegraphics[width=0.485\textwidth]{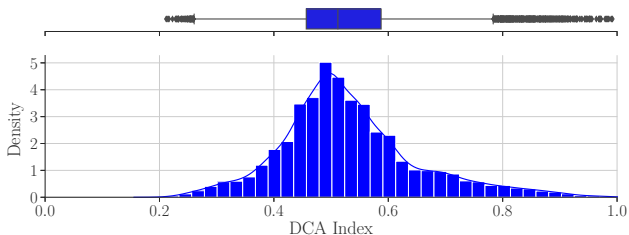}
\caption{Histogram and boxplot of the target variable.} \label{fig3}
\end{figure}

Exploratory data analysis began with a check of the main descriptive statistics of the model's input and output data. The input variables of the model follow a uniform distribution since they were sampled using LHS. The model's output variable, the DCA index, is a probability, and in this case, it ranges from a minimum value of 0.21 to a maximum value of 0.99. with a mean of 0.53 and a standard deviation of 0.12. Moreover, the mean and the median are almost the same (0.53 and 0.51), indicating a low number of outliers for this variable at the top of the distribution. A histogram and a boxplot were generated to visualize the distribution, as shown in Fig.~\ref{fig3}. The data of the target variable follows an approximately normal distribution around the average value.

Regarding the aspect and heading difference, these two angular variables are transformed into four numerical variables to calculate their respective sine and cosine values. Note that the variables referring to WEZ have minimum values of $-1$. These values are model adjustments for when it is not possible to estimate the WEZ.

\begin{comment}

\begin{table}
\centering
\caption{Descriptive statistics of the numerical model's input and output variables.}\label{tb4}
\begin{tabular}{|c|c|c|c|c|c|}
\cline{2-6}
\multicolumn{1}{c|}{} & 
\textbf{mean} & \textbf{std} & \textbf{min} & \textbf{median}  & \textbf{max} \\
\hline
distance [m]      & 54,676.38      & 25,828.54     & 3,564.59       & 46,653.23      & 92,556.43     \\
aspect [deg]                & 151.04 & 17.97 & 65.57 & 151.02 & 180.00 \\
delta\_head [deg]         & 106.68 & 50.50 & 0.03   & 116.80 & 179.99 \\
delta\_alt [ft] & 62.33         & 1,428.17      & -3,231.94       & 7.12       & 3,583.40      \\
delta\_vel [kn]       & 0.21   & 7.34  & -49.47   & 0.04    & 52.53  \\
wez\_max\_o2t [m]     & 24.04  & 10.90 & -1.00    & 27.34   & 40.11  \\
wez\_nez\_o2t [m]     & 8.22   & 1.45  & -1.00   & 8.36   & 12.44  \\
wez\_max\_t2o [NM]     & 20.37  & 15.41  & -1.00  & 27.31  & 41.06  \\
wez\_nez\_t2o [NM]     & 5.56   & 4.64  & -1.00  & 7.95   & 12.57  \\
vul\_thr\_bef\_shot  & 0.49   & 0.32  & 0.00   & 0.48 & 1.00   \\
vul\_thr\_aft\_shot    & 0.52   & 0.32  & 0.00   & 0.55   & 1.00   \\
shot\_point        & 0.56   & 0.32  & 0.00  & 0.63  & 1.00   \\
hp\_tgt\_off & 0.15   & 0.30  & 0.00   & 0.01   & 1.00   \\
hp\_thr\_vul & 0.12   & 0.27  & 0.00   & 0.01   & 0.99   \\
dca\_index           & 0.53   & 0.12  & 0.21    & 0.51  & 0.99   \\ \hline
\end{tabular}
\end{table}
\end{comment}

\subsection{Model Results}

The means of all 10-fold metrics, namely RMSE and $R^{2}$, used to evaluate the best model at the end of the grid search training process after performing the cross-validation are, respectively, $0.0543$ and 0.8020, while their standard deviations are 0.0009 and 0.0077. The coefficient of determination is approximately 80\%, and an RMSE, which penalizes the effects of outliers, is close to 0.05. Considering that this is a regression problem and that we are trying to predict the DCA index, which indicates a probability of success in this type of mission, the results are satisfactory for this type of problem since they would be making predictions with errors in the range of 5\% with a practically instantaneous inference time, which is desirable for a real-time application.

\begin{comment}

\begin{table}
\centering
\caption{Mean and standard deviation of the metrics used to evaluate the XGBoost model.}\label{tb3}
\begin{tabular}{|c|c|c|c|c|}
\cline{2-5}
\multicolumn{1}{c|}{} &
\textbf{MAE} & \textbf{MSE} & \textbf{RMSE} & \textbf{Coefficient of Determination ($R^{2}$)} \\
\hline
\textbf{mean}               
& 0.0400            
& 0.0030           
& 0.0543             
& 0.0802                                               \\
\textbf{std} 
& 0.0006            
& 0.0001            
& 0.0009             
& 0.0077    \\                             
\hline
\end{tabular}
\end{table}

\end{comment}

\begin{comment}

\begin{table}
\centering
\caption{Mean and standard deviation of the metrics used to evaluate the XGBoost model.}\label{tb3}
\begin{tabular}{|c|c|c|}
\cline{2-3}
\multicolumn{1}{c|}{} &
\textbf{RMSE} & \bm{$R^{2}$} \\
\hline
\textbf{mean}               
& 0.0543             
& 0.8020                                               \\
\textbf{std} 
& 0.0009             
& 0.0077    \\                             
\hline
\end{tabular}
\end{table}
\end{comment}

\section{Conclusions and Future Work}

This work presented a supervised machine learning model through the XGBoost library to develop an engagement decision support tool for BVR air combat in DCA missions. The model represented some of the primary dynamics of the BVR combat, allowing the analyst to evaluate the parameters that influence this combat. Additionally, we analyze in advance the performance of an engagement made by a pilot in air combat through the agent's DCA index, predicting the outcome of this confrontation. This kind of prediction may be used as an innovative decision support system for the pilots in this air combat modality concerning whether to engage an opponent or not. Although the index does not inform which is the better action to take, it measures the performance of the actions taken by the agent through the proposed modelings.

The modeling considered the characteristics of the aircraft and their armaments, along with the beliefs about the opposing aircraft. In addition, we also used the shot philosophy for the aircraft and the pilot's level of risk aversion. The model showed an $R^2$ of 0.802 and an RMSE of 0.054. Assuming the average values of the DCA index as 0.53 and standard deviation of 0.12, the results showed relatively consistent values and good predictive power of the DCA index. With this degree of confidence in the model, it is possible to predict future pilot's conditions, even with a few samples. The regression model could calculate the average values of the DCA index in each engagement in a coherent manner, providing quick answers to the results of this air combat phase that could provide the pilot with an improvement in his situational awareness in real-time. Thus, through simulation data, it is possible to improve the employment of the best operational tactics for each situation in the complex context of BVR air combat. Furthermore, it contributes by avoiding the incorrect and careless use of weapons, improving DCA mission effectiveness. Finally, since the pilot's survival is a determining factor, it could decrease the number of friendly aircraft lost in real-life BVR air combat.

Future work should move towards using a more significant number of simulations since we only analyzed 10,316 engagement cases due to the substantial computational cost to generate them. In addition, the search for more variables to define the agent state at the beginning of the engagement, or to do a feature engineering to generate new variables from the existing ones, could improve the performance of the proposed model. Also, comparing the results of XGBoost with other regression algorithms would be interesting to understand which supervised learning method best suits this type of problem. Finally, the conception of other operational metrics, like the DCA index, or the combination of several of them, to assess the level of performance in an engagement could contribute to bringing more information to the model.

\section*{Acknowledgments}

This work was supported by Finep (Reference nº 2824/20). Takashi Yoneyama is partially funded by CNPq -- National Research Council of Brazil through the grant 304134/2-18-0. 

\bibliographystyle{IEEEtran}
\bibliography{IEEEabrv,MLreferences}

\end{document}